\PassOptionsToPackage{unicode}{hyperref}
\PassOptionsToPackage{hyphens}{url}
\PassOptionsToPackage{dvipsnames,svgnames,x11names}{xcolor}
\documentclass[
  10pt,
  letterpaper,
]{article}
\usepackage{xcolor}
\usepackage[margin=1in]{geometry}
\usepackage{amsmath,amssymb}
\usepackage[square,numbers,sort&compress]{natbib}
\usepackage{iftex}
\ifPDFTeX
  \usepackage[T1]{fontenc}
  \usepackage[utf8]{inputenc}
  \usepackage{textcomp} 
\else 
  \usepackage{unicode-math} 
  \defaultfontfeatures{Scale=MatchLowercase}
  \defaultfontfeatures[\rmfamily]{Ligatures=TeX,Scale=1}
\fi
\usepackage{lmodern}
\ifPDFTeX\else
\fi
\IfFileExists{upquote.sty}{\usepackage{upquote}}{}
\IfFileExists{microtype.sty}{
  \usepackage[]{microtype}
  \UseMicrotypeSet[protrusion]{basicmath} 
}{}
\makeatletter
\@ifundefined{KOMAClassName}{
  \IfFileExists{parskip.sty}{%
    \usepackage{parskip}
  }{
    \setlength{\parindent}{0pt}
    \setlength{\parskip}{6pt plus 2pt minus 1pt}}
}{
  \KOMAoptions{parskip=half}}
\makeatother
\usepackage{color}
\usepackage{fancyvrb}

\DefineVerbatimEnvironment{Highlighting}{Verbatim}{commandchars=\\\{\}}
\newenvironment{Shaded}{}{}

\newcommand{\BuiltInTok}[1]{\textcolor[rgb]{0.00,0.50,0.00}{#1}}

\newcommand{\ExtensionTok}[1]{#1}

\newcommand{\NormalTok}[1]{#1}

\usepackage{longtable,booktabs,array}
\usepackage{calc} 
\usepackage{etoolbox}
\makeatletter
\patchcmd\longtable{\par}{\if@noskipsec\mbox{}\fi\par}{}{}
\makeatother
\IfFileExists{footnotehyper.sty}{\usepackage{footnotehyper}}{\usepackage{footnote}}
\makesavenoteenv{longtable}
\usepackage{graphicx}
\makeatletter
\newsavebox\pandoc@box
\newcommand*\pandocbounded[1]{
  \sbox\pandoc@box{#1}%
  \Gscale@div\@tempa{\textheight}{\dimexpr\ht\pandoc@box+\dp\pandoc@box\relax}%
  \Gscale@div\@tempb{\linewidth}{\wd\pandoc@box}%
  \ifdim\@tempb\p@<\@tempa\p@\let\@tempa\@tempb\fi
  \ifdim\@tempa\p@<\p@\scalebox{\@tempa}{\usebox\pandoc@box}%
  \else\usebox{\pandoc@box}%
  \fi%
}
\def\fps@figure{htbp}
\makeatother
\usepackage{microtype}
\usepackage{needspace}
\usepackage{booktabs}
\usepackage{array}
\usepackage{titlesec}
\usepackage{fancyhdr}
\titleformat{\section}{\Large\bfseries}{\thesection}{0.65em}{}
\titleformat{\subsection}{\large\bfseries}{\thesubsection}{0.6em}{}
\usepackage{bookmark}
\IfFileExists{xurl.sty}{\usepackage{xurl}}{} 
\hypersetup{
  pdftitle={Wide Learning: Learning to Reach Evidence},
  pdfauthor={Junzhou Chen},
  pdfsubject={Effective epistemic reach and learning to reach evidence},
  colorlinks=true,
  linkcolor={Maroon},
  filecolor={Maroon},
  citecolor={Blue},
  urlcolor={Blue},
  pdfcreator={LaTeX via pandoc}}

\title{Wide Learning: Learning to Reach Evidence}
\author{Junzhou Chen\\
\small School of Intelligent Systems Engineering, Sun Yat-sen University\\
\small Shenzhen 518107, China\\
\small \texttt{chenjunzhou@mail.sysu.edu.cn}}
\date{}

\begin{document}
\maketitle

\section*{Abstract}\label{abstract}
\addcontentsline{toc}{section}{Abstract}

Machine learning is usually evaluated after an evidence interface has
been fixed. A dataset, sensor suite, query language, action set, or
experimental protocol determines which observations can be obtained, and
learning is judged by what it extracts from them. We study a
complementary capability. A learner's state can determine which
evidence-generating experiments it can reliably realise under bounded
resources, even when primitive affordances remain fixed. We call this
learner-relative experiment family its \textbf{effective epistemic
reach}, and use \textbf{Wide Learning} for task-relevant
learning-induced changes in that family.
We formalise effective reach relative to learner state, deployment
budget, reliability threshold, and evaluation distribution. In a
controlled construction, two hidden worlds have exactly the same public
observation law. An informative diagnostic exists in a fixed
five-primitive substrate. Before calibration, one address attempt
realises it with probability at most \(2^{-10}=1/1024\), below a
pre-specified \(0.95\) threshold; after calibration, held-out
realisation is \(1\). Public-channel total variation is \(0\), whereas
the realised diagnostic has total variation \(1\), and sealed binary
risk moves from approximately \(1/2\) to \(0\). The construction
establishes that learning can change effective epistemic reach even when
primitive affordances and deployment resources are held fixed. It opens
a complementary evaluation
question for learning systems: not only what they infer from available
evidence, but what informative evidence experience teaches them to bring
within reach.

\section{1. Introduction}\label{introduction}

Most learning problems begin after a consequential choice has already
been made. A dataset has been collected. A sensor suite has been
installed. A query grammar has been specified. An action space has been
exposed. A laboratory protocol has been made executable. Once this
evidence interface is fixed, the central questions are familiar: how
efficiently can a learner fit, compress, represent, generalise, plan, or
reason from the observations it receives? Progress on these questions
has transformed machine learning.

The fixed interface is often a useful abstraction. It separates
acquisition from inference, makes systems comparable, and gives
statistical learning theory a stable observation model. Yet it also
hides a capability difference. Two learners with the same primitive
tools and the same deployment budget need not be able to bring the same
task-relevant evidence into effective use. One may know how to
instantiate a diagnostic procedure, configure a sensor, compose a
sequence of actions, construct a query, or validate an assay that the
other cannot reliably realise. Their difference is not exhausted by how
well they process a common observation. It concerns which observations
they can make available in the first place.

Evaluation can therefore reverse the usual order. Instead of fixing the
evidence boundary and asking only what learning can extract within it,
we ask whether learning can change the boundary of effectively
accessible evidence. The contrast is:

\[
\boxed{\text{learning from evidence}\quad\longleftrightarrow\quad\text{learning to reach evidence}.}
\]

The relation to the familiar term \textbf{Deep Learning} is functional
rather than architectural. \textbf{Deepening} improves extraction from
the current effective evidence envelope, whereas \textbf{widening}
changes that envelope by learning how to realise new task-relevant
experiments. These are complementary capability coordinates, not
mutually exclusive algorithm classes. A neural model may support either
or both, and ``wide'' here does not refer to parameter width or to a
rival network architecture.

The phrase \textbf{effective epistemic reach} names this missing
coordinate: the experiments a learner can reliably bring to bear under
specified resources. Logical availability alone is insufficient. An
experiment found only through astronomically unreliable search differs,
under a bounded deployment criterion, from one that can be instantiated
on demand. The quantity is learner-relative because learned states can
support different experiment families on the same substrate, and
task-sensitive because only relevant changes in observation law or
decision risk count.

The phrase ``wide learning'' has appeared previously in unrelated
senses, including infinite-width learning, automated feature
engineering, and Fujitsu's Wide Learning™ explainable-AI technology
\citep{pandey2014deepwide,banerjee2016wide,fujitsu2018wide}. No lexical priority is claimed. Here the term refers
exclusively to task-relevant, learning-induced change in effective
epistemic reach.

Several established traditions provide the conceptual foundations.
Statistical
comparison of experiments supplies a language for ordering information
structures \citep{blackwell1953experiments}. Bayesian experimental design treats experiments as
decision variables \citep{lindley1956measure}. Rational inattention makes information
processing and acquisition costly \citep{sims2003inattention}. Active learning and active
perception let a learner choose data or sensing actions \citep{cohn1994active,bajcsy1988active}.
Reinforcement learning studies information-seeking behaviour \citep{bellemare2016count}.
Options and affordances describe extended or relational action
possibilities \citep{sutton1999options,gibson1979ecological}, while meta-learning changes adaptation itself
\citep{finn2017maml}. Automated science closes loops from hypothesis to experiment
\citep{king2009automation}, and continual-learning research studies whether systems retain
the capacity to adapt \citep{dohare2024plasticity}. We treat learning-induced change in
effective experiment capability as a cross-domain object, with a common
budgeted definition and a direct distinguishability test.

A valid test holds primitive affordances fixed, rules out information
from post-processing an exactly uninformative public channel, and links
the learned procedure to a task-relevant observation or decision. A
learned program qualifies only when it makes an informative experiment
effectively realisable under the stated criterion. These controls make
the capability change falsifiable.

The controlled construction fixes five requirements. Primitive
affordances remain unchanged, and the public evidence is exactly
insufficient. An informative diagnostic lies below the
effective-realisability threshold under a fixed deployment budget.
Learning makes that diagnostic constructible on fresh instances, and the
resulting observation changes a sealed downstream decision. Together,
these requirements isolate the transition.

The construction uses paired worlds that share a public descriptor and
public response law but differ in a hidden binary mechanism. A valid
diagnostic address is generated by an unknown affine map over
\(\mathbb F_2^{10}\). Before calibration, one address attempt succeeds
with probability at most \(1/1024\), far below the pre-specified
\(0.95\) effective-realisability threshold. Calibration exposes 16
descriptor--address pairs through the same self-testing primitive. A
reference learner recovers the affine map by exact Gaussian elimination
and synthesises valid addresses for fresh descriptors with probability
one. The valid probe separates the paired worlds and eliminates sealed
decision error. Public-only, generic-instrumentation, fixed-repertoire,
sham, memorisation, compute, leakage, and plasticity controls rule out
the intended shortcuts.

We make three contributions. First, we distinguish learning that
improves inference on available evidence from learning that changes
which evidence-generating experiments are effectively accessible.
Second, we formalise effective
epistemic reach as a learner-, budget-, reliability-, and
distribution-relative experiment family, and state the elementary
post-processing boundary that makes exact public equivalence
consequential. Third, we provide a controlled existence test in which
learned state, rather than a new primitive, moves an informative
diagnostic across a pre-specified reach threshold and changes both
paired-world distinguishability and sealed decision risk.

\section{2. Epistemic Reach}\label{epistemic-reach}

\subsection{2.1 Effective epistemic
reach}\label{effective-epistemic-reach}

Let \(w\in\mathcal W\) denote a possible world and \(S\) the current
state of a learner. A world-directed procedure induces an observation
law on a relevant outcome space. Let \(e\in\mathcal E\) denote the
resulting statistical-experiment equivalence class, with world-indexed
law \(P_w^e\). Such a class may be realised by a physical measurement,
an intervention, a software query, a sequence of actions, a diagnostic
program, or a validated composition of primitive operations.

Two procedures are treated as realisations of the same statistical
experiment when they induce the same world-indexed observation law on
the relevant outcome space. Effective reach therefore concerns
experiment equivalence classes rather than API names, program syntax,
option boundaries, or a particular decomposition into primitive actions.
A target experiment family belongs to effective reach when at least one
admissible realisation strategy meets the fixed resource and reliability
criterion over the evaluation distribution.

Let \(B\) collect the deployment constraints relevant to realisation,
including time, queries, action steps, energy, memory, and permitted
primitives. Let \(\xi\sim\mu\) index an instance drawn from a fixed
evaluation or meta-environment distribution. An instance-indexed
\textbf{target experiment family} is a map

\[
\boxed{E:\Xi\to\mathcal E,\qquad \xi\mapsto E_\xi.}
\]

Let \(\pi_S\) denote the experiment-realisation policy or constructor
induced by learner state \(S\) under the fixed primitive substrate. For
learner state \(S\), define the realisation probability

\[
\rho_{B,\mu}(S,E)
=
\Pr_{\xi\sim\mu,\,\omega}
\!\left[\pi_S(\xi,\omega)\text{ realises }E_\xi\text{ within }B\right],
\]

where \(\omega\) collects learner randomisation and observation noise
when either is present. For a deterministic noiseless evaluation, the
probability is only over \(\xi\). A fixed target experiment
\(e^\star\in\mathcal E\) is recovered by taking the target map to be
constant, \(E_\xi\equiv e^\star\) for all \(\xi\in\Xi\). Given a
permitted failure rate \(\eta\), the
\textbf{effective epistemic reach} is

\[
\boxed{
\mathfrak R_{B,\eta,\mu}(S)
=
\left\{E:\Xi\to\mathcal E\,\middle|\,\rho_{B,\mu}(S,E)\ge 1-\eta\right\}.
}
\]

The policy \(\pi_S\) may arise through supervised learning,
reinforcement learning, meta-learning, program synthesis, or other
mechanisms. Effective epistemic reach describes the capability induced
by the resulting learner state, independently of the learning machinery.

The realisation profile \(\rho_{B,\mu}(S,E)\) is the underlying graded
quantity. The thresholded family \(\mathfrak R_{B,\eta,\mu}(S)\) is an
operational slice induced by a pre-specified evaluation criterion rather
than a unique intrinsic boundary. Varying \(B\) or \(\eta\) yields a
reliability--resource profile. The controlled construction uses a fixed
threshold only to make the existence test falsifiable and reproducible.

The family \(\mathfrak R_{B,\eta,\mu}(S)\) is defined at the
distribution level: its members satisfy an average realisation criterion
under \(\mu\). Exact observational equivalence makes the stronger
comparison between target experiments themselves. Membership under the
average criterion leaves open whether every \(E_\xi\) is realised on
every instance.

Membership is operational. It requires reliable instantiation and
execution under the fixed protocol; the mere existence of a finite
program somewhere in the substrate is insufficient. When \(B\), \(\eta\), and \(\mu\)
are fixed, we write \(\mathfrak R(S)\) for brevity. The controlled
construction fixes \(1-\eta=0.95\), a one-attempt deployment budget, and
a distribution over independently sampled affine meta-environments and
held-out descriptors.

The definition separates three objects that are often conflated. The
\textbf{primitive substrate} specifies atomic calls available to every
compared learner. The learner state \(S\) contains acquired parameters,
procedures, or skills. The reach family \(\mathfrak R_{B,\eta,\mu}(S)\)
is the resulting set of target experiment families the learner can
instantiate at the required reliability. A fixed substrate therefore
does not imply a fixed effective experiment family. It fixes the alphabet,
not every bounded composition the learner can reliably construct.

Predictor adaptation under fixed evidence may improve predictions while
leaving epistemic reach unchanged. Reach changes only when the update
alters which observation law can be induced. The newly realisable
procedure must also be epistemically relevant: for some task-relevant
worlds, it changes the available observation distributions or lowers the
best attainable decision risk.

\subsection{2.2 Observational
equivalence}\label{observational-equivalence}

Exact observational equivalence remains a useful general theoretical
object. For fixed \(B,\eta,\mu\), define

\[
w\sim_{S,B,\eta,\mu}w'
\quad\Longleftrightarrow\quad
P_w^{E_\xi}=P_{w'}^{E_\xi}\;\text{ for every }
E\in\mathfrak R_{B,\eta,\mu}(S)\text{ and }\mu\text{-almost every }\xi.
\]

The relation says that every instance-specific experiment named by the
stated effective target families induces the same law in the two worlds.
It remains relative to \(S\), \(B\), \(\eta\), and \(\mu\), and therefore
leaves open distinctions available to another learner, instrument, or
resource regime.

The controlled construction proves exact equivalence for the public
channel and uses a pre-specified reliability threshold to place the
task-relevant diagnostic outside pre-learning effective reach. Other
executable pre-learning instrumentation procedures are assessed by that
threshold, not by an assertion of exact full-envelope equality.

At the experiment-family level, the question connects directly to statistical decision
theory. Blackwell comparison asks when one statistical experiment is at
least as informative as another for every decision problem \citep{blackwell1953experiments}.
Wide Learning asks how learning changes the experiment envelope available
to a particular bounded learner. Most real experiment
families are only partially ordered, so this existence result requires
neither a universal scalar score nor global dominance. A single
task-relevant split, paired with a sealed risk consequence, is enough
for an existence result.

\subsection{2.3 Fixed-meta-interface
reduction}\label{fixed-meta-interface-reduction}

Any finite system can be redescribed as operating behind a sufficiently
broad fixed meta-interface. A robot that learns a new sensing routine
still uses its motors and sensors; a program that constructs a query
still uses a fixed instruction set; a laboratory controller that learns
a protocol still invokes a finite set of devices. Under this description,
the same capability difference appears as the ability to compile an
experiment from fixed primitives.

Under a fixed-meta-interface description, the relevant variable is
whether the learner can compile task-relevant experiments from the
primitive substrate within \(B\). If one state can reliably instantiate
a diagnostic and another cannot, their effective reach differs even
though both computations are expressible in the same universal language.
Treating all logically expressible programs as equally available would
erase resource-bounded distinctions throughout computer science.
Effective reach retains the distinction that deployment makes
operational.

The representation is therefore intentionally neutral between ``new
interface'' and ``new skill'' descriptions. We hold the substrate fixed
when comparing learner states, specify the deployment budget, and ask
whether the induced experiment family changes. This reduction blocks the
trivial strategy of declaring every new learned procedure a new sensor,
while also blocking the opposite strategy of making every procedure free
by placing it inside an unbounded meta-interface.

\section{3. Learning to Reach
Evidence}\label{learning-to-reach-evidence}

\subsection{3.1 The Wide transition}\label{the-wide-transition}

Consider learning that updates the state

\[
S_0\longrightarrow S_1.
\]

A \textbf{Wide transition} for a fixed task, budget, reliability
threshold, and evaluation distribution occurs when a task-relevant
target experiment family \(E^\star:\Xi\to\mathcal E\) moves from outside
to inside effective reach:

\[
E^\star\notin\mathfrak R_{B,\eta,\mu}(S_0),
\qquad
E^\star\in\mathfrak R_{B,\eta,\mu}(S_1),
\]

and this change separates at least one task-relevant pair of worlds or
strictly lowers the attainable sealed decision risk. This deliberately
minimal criterion allows capability trade-offs and places no requirement
on global reach inclusion, new primitives, or algorithmic novelty.

The distinction between logical and effective reach is essential. Before
learning, each \(E^\star_\xi\) may admit one address among many, one long
program among many, or one protocol among many. A nonzero accidental
discovery probability does not make it effectively realisable under a
reliability criterion. The scientific statement must therefore report
the budget, success probability, and threshold. ``Outside effective
reach under \(B\)'' is a bounded capability claim; ``logically
impossible'' would be a different and much stronger claim.

\subsection{3.2 A post-processing
boundary}\label{a-post-processing-boundary}

The importance of reach follows from an elementary consequence of
identical observation laws.

\textbf{Post-processing boundary (standard).} Let
\(w_0,w_1\in\mathcal W\). If the observation laws induced by the
currently effective experiment family are equal in the two worlds, then
no world-independent deterministic or randomised post-processing of
those observations can distinguish \(w_0\) from \(w_1\).

\textbf{Proof.} Fix any \(E\in\mathfrak R_{B,\eta,\mu}(S)\) and relevant \(\xi\),
and let the common observation law of \(E_\xi\) be \(P\). Any
world-independent post-processing rule is
a Markov kernel \(K\) from observations to outputs. The induced output
law is \(KP\) in both worlds. Mixtures, adaptive combinations whose
complete trace laws are equal, and arbitrarily expensive internal
computation are all special cases of such kernels. Therefore the output
laws remain equal. \(\square\)

Stronger reasoning over an unchanged observation law cannot substitute
for an experiment that changes that law. In the controlled construction,
the boundary applies exactly to the public channel; the diagnostic's
crossing of the fixed reliability threshold establishes the reach
transition separately.

\subsection{3.3 Complementary capability
coordinates}\label{complementary-capability-coordinates}

For the functional contrast used here, deepening improves the map from
the current effective evidence envelope to representation, prediction,
or action, whereas widening is a learning-induced change in that
envelope. These are capability coordinates, not mutually exclusive
algorithm classes. A learned experiment constructor may itself be
represented by a neural network; better representation may be required
to recognise when a probe is valid; and a wider evidence envelope may
supply data that enables deeper inference.

Active learning already selects informative labels, active sensing
controls viewpoints, and reinforcement learning values epistemic
actions. Wide transitions may arise through learned selection,
composition, control, synthesis, or other mechanisms whenever learning
changes which task-relevant experiment families meet the
effective-realisability criterion. A common reach variable makes this
change measurable across domains.

Figure 1 summarises the target transition. Before learning, paired
worlds share the public law and the informative diagnostic is below the
pre-specified effective-realisability threshold. Calibration changes
learner state while the primitive substrate remains fixed. After
learning, the diagnostic is constructed within the same deployment
budget, the paired worlds induce different probe outcomes, and the
sealed decision becomes solvable.

\begin{figure}[!htbp]
\centering
\includegraphics[width=0.80\linewidth]{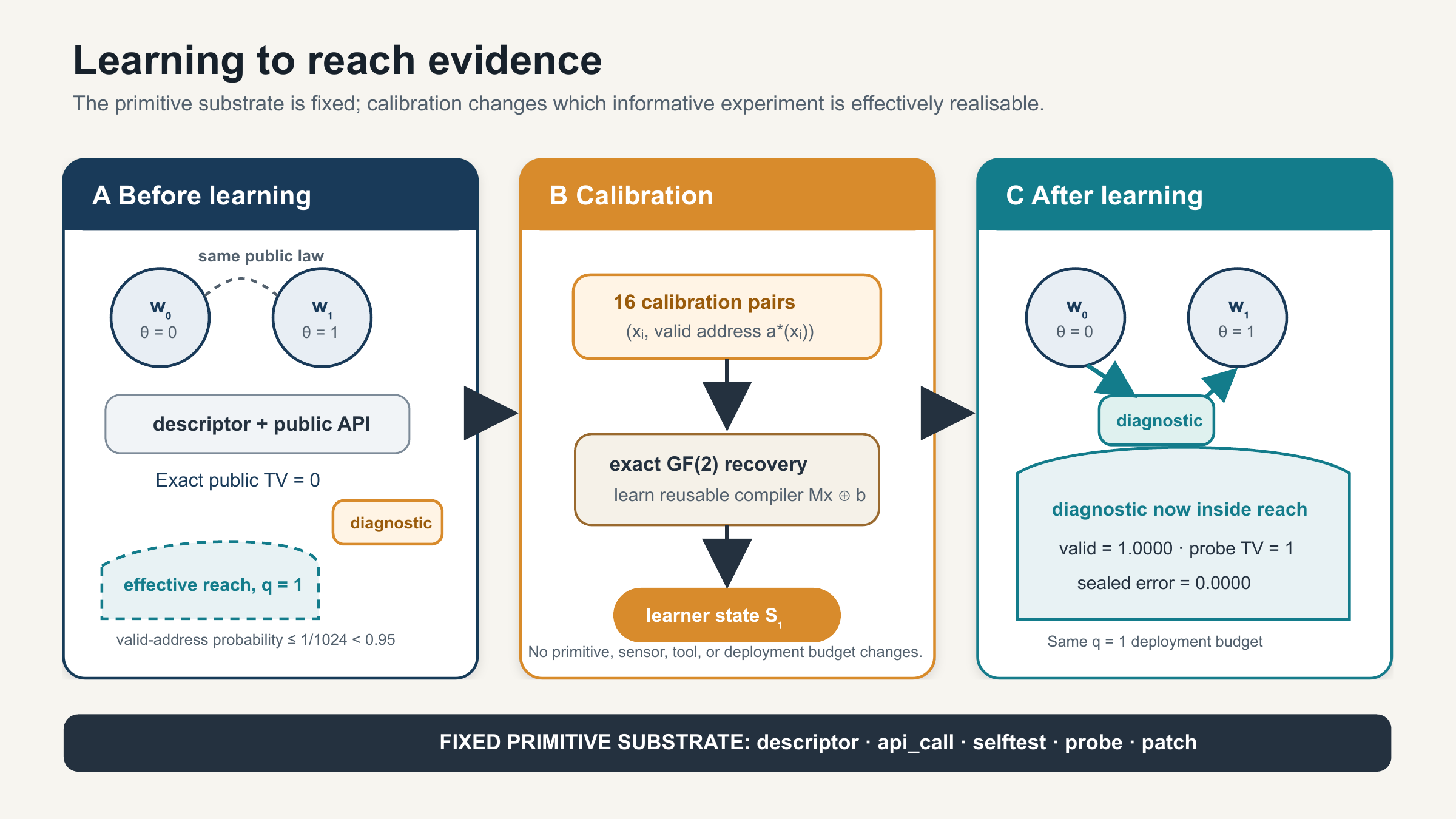}
\caption{Deep improves what can be extracted within current epistemic
reach. Wide Learning changes what evidence enters that reach. In the
controlled construction, the primitive substrate is fixed; learning
changes the learner's ability to realise the informative diagnostic
under the same deployment budget. The dashed boundary in panel A denotes
effective reach under the fixed reliability criterion, not logical
impossibility. Calibration changes the learner state but adds no
primitive affordance. Panel C shows a task-relevant split: the learned
diagnostic enters effective reach, returns different outcomes in the
paired worlds, and determines the sealed patch.}
\end{figure}

\section{4. A Controlled Existence
Test}\label{a-controlled-existence-test}

\subsection{4.1 Construction and evaluation
protocol}\label{construction-and-evaluation-protocol}

We use a deliberately minimal controlled construction to isolate the
transition. The setting is noiseless and exact. Its parameterisation,
comparison arms, controls, and \(0.95\) reliability criterion were fixed
before the reference run.

Each meta-environment contains paired worlds

\[
w_0=(x,\theta=0),\qquad w_1=(x,\theta=1),
\]

\Needspace{7\baselineskip}

where \(x\in\mathbb F_2^{10}\) is public and \(\theta\) is the hidden
mechanism relevant to a final binary patch. Every paired world exposes
the same five primitives:

\[
\texttt{descriptor},\quad\texttt{api\_call},\quad\texttt{selftest},\quad\texttt{probe},\quad\texttt{patch}.
\]

The descriptor returns only \(x\). The public API may respond as a
function of \(x\), the query, and the adaptive public history, but its
code and response law are exactly independent of \(\theta\). The
self-test reports whether a proposed address is valid and reveals
nothing about \(\theta\). The probe returns \(\theta\) only at the valid
address and a fixed null symbol elsewhere. The patch commits a final
binary decision; only after commitment may the evaluator consult the
hidden truth. These primitives and their semantics are identical before
and after learning.

For each of 20 independently seeded meta-environments, an invertible
matrix \(M\in GL(10,2)\) and bias \(b\in\mathbb F_2^{10}\) are sampled.
The valid diagnostic address for descriptor \(x\) is

\[
a^\star(x)=Mx\oplus b.
\]

Because the map is affine and invertible, each fresh descriptor has one
valid address in a space of \(2^{10}=1024\) addresses. The deployment
budget \(B_{\rm dep}\) permits at most one address attempt. The
effective-realisability criterion requires valid construction
probability at least \(0.95\).

The public law and the instrumentation claim are kept separate. The
public channel is exactly equivalent across the paired worlds:
\(TV_{\mathrm{public}}=0\). Under an unknown random affine map, any
\(\theta\)-independent one-address policy has ensemble hit probability
bounded by

\[
P(\text{valid diagnostic})\le \frac{1}{2^{10}}=\frac{1}{1024}=0.0009765625.
\]

The bound follows because for any fixed public descriptor and any
address selected without knowledge of the random bias, \(a^\star(x)\) is
uniform over the address space. Under the fixed evaluation distribution
and threshold,

\[
\rho_{B_{\rm dep},\mu}(S_0,E^\star)\le 2^{-10}=1/1024<0.95,
\qquad
\rho_{B_{\rm dep},\mu}(S_1,E^\star)=1.
\]

Therefore
\(E^\star\notin\mathfrak R_{B_{\rm dep},\eta,\mu}(S_0)\) and
\(E^\star\in\mathfrak R_{B_{\rm dep},\eta,\mu}(S_1)\). The pre-learning
probability is nonzero, and finite samples can produce a hit rate above
its ensemble value. The licensed claim is that the diagnostic is below
the effective-realisability threshold before calibration and perfectly
realisable after calibration.

\subsection{4.2 Calibration and learned experiment
construction}\label{calibration-and-learned-experiment-construction}

Calibration provides 16 pairs \((x_i,a^\star(x_i))\). The addresses are
obtained through allowed self-testing, not by exposing \(M\), \(b\),
\(\theta\), or held-out answers. The calibration descriptors are
selected so that the augmented design has rank 11 and therefore
identifies the affine map. The reference learner solves the resulting
linear systems by Gaussian elimination over \(\mathbb F_2\), yielding a
rule that synthesises addresses for held-out descriptors.

Calibration and deployment use separate budgets. The learning budget
\(B_{\rm cal}\) supports calibration, whereas effective reach is
evaluated under the fixed deployment budget \(B_{\rm dep}\). The
construction converts calibration-time search into a reusable
deployment-time experiment-realisation capability.

At deployment, the learned state performs a fixed sequence: read a fresh
\(x\); compute \(\hat a(x)\); self-test that address once; probe once if
valid; and patch according to the probe. The capability acquired during
calibration is a reusable experiment-construction rule. No correct
address or task-specific menu is supplied at deployment, and the address
search budget remains one.

The comparison includes four core arms. The public-only arm receives the
same public evidence and at least as many nominal public-compute
iterations as the reference learner, but makes no instrumentation call.
The pre-learning instrumentation arm has the full primitive interface
but no learned affine map and may attempt one generic address. The
post-learning arm uses the recovered affine map under the same
deployment budget. The oracle receives the correct address but not
\(\theta\), establishing the achievable decision floor once the
diagnostic is reachable.

The analytic bound and exact affine recovery are the primary evidence.
The reference implementation evaluates each arm on 20 meta-environments,
64 fresh held-out descriptors per environment, and both hidden
mechanisms per descriptor, yielding 2,560 sealed worlds per arm. These
runs provide executable verification, finite-run confirmation, and
reproducibility records. Table 1 reports their endpoints alongside the
analytic ensemble probability.

\begingroup\small

{\def\LTcaptype{none} 
\begin{longtable}[]{@{}
  >{\raggedright\arraybackslash}p{(\linewidth - 8\tabcolsep) * \real{0.2000}}
  >{\raggedleft\arraybackslash}p{(\linewidth - 8\tabcolsep) * \real{0.2000}}
  >{\raggedleft\arraybackslash}p{(\linewidth - 8\tabcolsep) * \real{0.2000}}
  >{\raggedleft\arraybackslash}p{(\linewidth - 8\tabcolsep) * \real{0.2000}}
  >{\raggedleft\arraybackslash}p{(\linewidth - 8\tabcolsep) * \real{0.2000}}@{}}
\toprule\noalign{}
\begin{minipage}[b]{\linewidth}\raggedright
Condition
\end{minipage} & \begin{minipage}[b]{\linewidth}\raggedleft
Diagnostic realisation probability
\end{minipage} & \begin{minipage}[b]{\linewidth}\raggedleft
Public-channel TV
\end{minipage} & \begin{minipage}[b]{\linewidth}\raggedleft
Valid-probe TV
\end{minipage} & \begin{minipage}[b]{\linewidth}\raggedleft
Sealed error
\end{minipage} \\
\midrule\noalign{}
\endhead
\bottomrule\noalign{}
\endlastfoot
Public only & 0 & 0 & N/A & 0.5000 \\
Pre-learning instrumentation & 0.0015625 observed; \(1/1024\) ensemble &
0 & 1 if valid & 0.49921875 \\
Post-learning & 1.0000 & 0 & 1 & 0.0000 \\
Oracle & 1.0000 & 0 & 1 & 0.0000 \\
\end{longtable}
}

\endgroup

\textbf{Table 1 \textbar{} Controlled-construction endpoints.}
Public-channel total variation is exactly zero in every condition.
Valid-probe total variation is conditional on realising the valid
diagnostic. The central transition is in diagnostic realisation
probability: the pre-learning finite-run rate is \(0.0015625\), while
the ensemble probability is at most \(1/1024\); both remain far below
the \(0.95\) criterion. Post-learning and oracle realisation
probabilities are one.

The reference transition crosses the pre-specified reach threshold:
valid-diagnostic synthesis rises from \(0.0015625\) to \(1.0\). Public
observations remain identical for \(\theta=0\) and \(\theta=1\), while a
valid learned probe returns a point mass at 0 in one world and a point
mass at 1 in the other, giving total variation 1. The permissive
pre-learning instrumentation arm has error \(0.49921875\), close to the
balanced chance floor; the post-learning arm has error 0.

\subsection{4.3 Alternative-explanation
controls}\label{alternative-explanation-controls}

Eight controls target the shortest alternative explanations. First,
public equivalence is certified structurally: the public callables are
parsed as abstract syntax trees and contain no read or branch on
\(\theta\); exhaustive paired traces over all 1,024 descriptors in each
meta-environment have zero mismatches. Second, the one-attempt reach
bound is analytic, with Monte Carlo treated only as confirmation. Third,
descriptor leakage is excluded by construction because every held-out
\(x\) appears with both values of \(\theta\).

Fourth, a fixed-repertoire learner may reuse calibration addresses but
may not infer the affine rule. It tests whether success can be explained
by lookup from an experimenter-supplied, test-specific menu, while
leaving learned selection in general open. Since an invertible affine
map is bijective and held-out descriptors are fresh, its valid rate is
exactly zero. Fifth, sham probes differ from the valid address and always return
the same null symbol in both worlds; their sealed error is exactly
\(0.5\). Sixth, memorisation is tested with fresh descriptors and 20
independently sampled affine maps; training--held-out overlap is zero
and post-learning valid synthesis is 1.0 for every meta-seed.

Seventh, the public-only and generic-instrumentation controls receive
4,096 nominal public-compute iterations, compared with 100
multiply--accumulate positions for the reference affine evaluation.
Exact public equivalence makes post-processing insufficient to reveal
\(\theta\), independently of the nominal compute allowance. The operation
counts provide a matched-compute control, not a wall-clock equivalence.
Eighth, a separate fixed-evidence parity task verifies that ordinary
adaptation can reduce prediction error from \(0.5\) to 0 without any
instrumentation call. That success is recorded as generic plasticity,
not reach expansion.

All eight controls pass, as do the ten corresponding validity checks for
public leakage, sufficient search, extra affordance, externally supplied
test-specific menus,
compute-only explanation, validator or label leakage, memorisation,
generic plasticity, epistemically irrelevant tooling, and absent
decision consequence. The reference implementation retains the complete
code-level certificates and per-world records.

\section{5. Interpretation and Relation to Existing
Work}\label{interpretation-and-relation-to-existing-work}

\subsection{5.1 What the construction
establishes}\label{what-the-construction-establishes}

The controlled construction establishes an existence claim with three
linked parts. First, effective experiment capability is not fixed by the
primitive substrate alone. The five calls available at deployment are
identical in \(S_0\) and \(S_1\), yet the learner's probability of
constructing the valid diagnostic on fresh descriptors crosses from
below threshold to one. Second, this is an epistemic change rather than
behavioural novelty: the hidden pair has exact public equivalence, while
the realised diagnostic produces disjoint outcomes. Third, the new
distinction has a sealed consequence: it removes the balanced binary
decision error.

The primitive substrate and probe semantics are identical before and
after calibration. The informative address is already present; learning
changes the learner's ability to construct it within one attempt. Exact
public equivalence also isolates this capability change from additional
reasoning over the old channel. World-independent computation preserves
the equality of its laws, whereas the learned experiment supplies a
different observation.

Recovering the affine map is deliberately conventional. Its role is to
separate algorithmic novelty from the capability under study: an
ordinary learning update changes which informative experiment can be
reliably instantiated at deployment.

A fixed-interface evaluation assigns the two learner states the same
Bayes-optimal performance on the public channel:

\[
R_{\mathrm{public}}(S_0)=R_{\mathrm{public}}(S_1)=\frac12,
\]

because the public law is exactly independent of \(\theta\). A
reach-aware evaluation separates them immediately:

\[
\rho_{B_{\rm dep},\mu}(S_0,E^\star)\le\frac{1}{1024},
\qquad
\rho_{B_{\rm dep},\mu}(S_1,E^\star)=1.
\]

Thus

\[
\boxed{\text{fixed-interface performance equivalence}\ \not\Rightarrow\ \text{effective-reach equivalence}.}
\]

\subsection{5.2 Relation to neighbouring
traditions}\label{relation-to-neighbouring-traditions}

Statistical experiment theory provides the mathematical substrate.
Blackwell comparison orders experiments by their value across decision
problems \citep{blackwell1953experiments}; effective epistemic reach tracks how the
bounded experiment family changes with learner state. Related work on
epiplexity studies the structure a bounded observer can extract from a
data source \citep{finzi2026epiplexity}, while state-dependent teaching models show how
acquired prerequisites change the interpretation of supplied signals
\citep{taskesen2026understanding}. Effective reach locates the corresponding learner dependence
in world-directed experiment realisation.

Experimental design and information-acquisition theory supply objectives
and costs for choosing evidence. Lindley's criterion values experiments
by expected information \citep{lindley1956measure}; rational inattention and modern cost
theories formalise limits on acquired or processed information
\citep{sims2003inattention,denti2022experimental,pomatto2023cost}. Machine learning already
constructs or amortises informative designs, including learned
behavioural experiments and Deep Adaptive Design
\citep{valentin2024behavioral,foster2021dad}. Within this tradition, effective reach makes the
before-and-after deployment capability itself the measured learner
variable.

Active learning, active perception, and active diagnosis choose which
labels, viewpoints, or observations to obtain
\citep{cohn1994active,bajcsy1988active,pu2017acquire}. Recent systems extend this lineage
through meta-learned feature acquisition, in-context query policies, and
test-time information gathering
\citep{kobayashi2026learning,russo2026icpe,cooper2025curious}. Adaptive-camera control further
shows how deployment-time sensor adjustment can improve the observation
presented to a model, and adaptive sensing has been proposed as a
first-class AI agenda \citep{baek2025camera,baek2025sensing}. A Wide transition occurs when
learning through any of these mechanisms changes which task-relevant
experiment families meet the stated resource and reliability criterion.

Reinforcement learning, skill acquisition, and affordance theory explain
how information-gathering action becomes executable. Epistemic
exploration rewards uncertainty-reducing visits \citep{bellemare2016count}; options and
Gibsonian affordances describe extended and learner-relative action
possibilities \citep{sutton1999options,gibson1979ecological}. Meta-learning can acquire
representations or policies that support rapid task adaptation, including
safe acquisition functions learned across systems
\citep{finn2017maml,schrum2022metaactive}. These mechanisms widen reach when the acquired
policy, option, or adaptation rule brings an informative trajectory
across the effective-realisability threshold.

Automated science combines hypothesis formation, experiment selection,
physical execution, and interpretation, as demonstrated by Robot
Scientist Adam \citep{king2009automation}. Continual-learning plasticity addresses whether
the capacity to adapt survives accumulated experience
\citep{dohare2024plasticity}. These settings expose complementary aspects of the problem:
autonomous systems provide rich experiment families, while plasticity
supports continued acquisition of the procedures that realise them.
Effective reach measures the epistemic consequence of that acquisition
on a specified task.

Taken together, these traditions provide mechanisms for selecting,
interpreting, and executing information-gathering actions. The
complementary question is whether learning changes the set of
world-directed experiments that a system can reliably realise under a
fixed deployment budget. Effective epistemic reach makes this question
explicit and comparable across statistical decision models, sensors,
queries, interventions, software diagnosis, and laboratory actions. The
controlled construction isolates this capability change while holding
primitive affordances fixed.

\subsection{5.3 Scope and limitations}\label{scope-and-limitations}

The evidence comes from a noiseless constructed environment. Calibration
uses exhaustive self-testing, the affine family is identifiable from 16
selected descriptors, and the reference learner has the correct
hypothesis class. The construction therefore establishes existence under
these conditions, with no result on noise robustness, sample complexity,
scaling, or naturalistic prevalence. The observed pre-learning hit rate,
four of 2,560 worlds or \(0.0015625\), is slightly above the \(1/1024\)
ensemble expectation but remains far below the reach threshold. The
matched-compute counts are nominal; exact public equivalence carries the
post-processing argument. More generally, reach is task- and
budget-specific: current evidence may already suffice, and additional
experiments may be too costly, unsafe, or invalid.

\section{6. Toward a Science of Epistemic
Reach}\label{toward-a-science-of-epistemic-reach}

An existence result becomes useful when it changes what researchers
measure. Current evaluations typically condition on a fixed dataset,
toolset, sensor suite, or environment API. A reach-aware evaluation
would additionally hold primitive affordances and deployment resources
fixed, expose calibration experience, and ask whether the learner
acquires reusable procedures that induce new task-relevant observation
laws on held-out instances. The primary endpoints would be experiment
realisation, equivalence splitting, and sealed decision risk, not task
accuracy alone.

The most immediate question is resource allocation. When should a
learner spend computation extracting more from current evidence, and
when should it invest in acquiring a procedure that changes what
evidence can be obtained? The answer requires costs on construction,
validation, and deployment as well as value of information. Existing
decision theory and rational-inattention tools are natural starting
points; the empirical variable is the learned reach profile to which
those tools are applied.

A second question is measurement in naturalistic systems. Real
experiments are noisy, partially valid, and difficult to enumerate.
Reach will need operational proxies: held-out protocol synthesis
success, statistically powered equivalence splitting, calibrated
confidence in instrument validity, or task-relative Bayes-risk reduction
under matched resources. Evaluations must distinguish learning a
genuine acquisition capability from trivial lookup, leakage, or an
externally supplied test-specific menu.

A third question concerns epistemic relevance. Systems continually
acquire behavioural skills, tools, and routines. Which ones are
informative rather than merely novel? The clean criterion is
counterfactual: does the acquired procedure induce an observation law
that separates task-relevant possibilities which remained equivalent
throughout the previous effective envelope? Approximate versions can use
total variation, test power, or decision value, but the task and
threshold should be fixed in advance.

Scaling is another open empirical axis. Reach might expand with data,
model capacity, curriculum diversity, embodied experience, or time; it
might also contract as systems lose plasticity or over-specialise. The
present endpoint allows future scaling studies to separate better
prediction from changed experiment capability.

Finally, systems with similar current performance may have different
future capacities to widen. One may have learned reusable experimental
abstractions while another has only fitted present observations.
Detecting that difference could matter for continual deployment,
scientific automation, and long-horizon agency. The current reach
variable provides a concrete target for future theories of that capacity.

Together, these questions define a measurement programme. The immediate
empirical priorities are independent verification of the controlled
construction and naturally occurring instances in which substrate,
budget, validity, and task consequence can be fixed with comparable
precision.

\section{7. Conclusion}\label{conclusion}

Machine learning has become exceptionally good at extracting structure
after evidence is made available. That success can obscure a separate
capability: learning can sometimes change which informative evidence a
bounded system can bring within effective reach.

We formalised this capability through a learner-relative experiment
family and an induced observational equivalence relation. The formal
boundary is elementary but decisive: world-independent post-processing
cannot separate worlds that induce the same laws throughout the current
effective experiment envelope. The controlled construction supplies a
minimal witness. Under a fixed five-primitive substrate and one-attempt
deployment budget, calibration teaches an affine diagnostic constructor.
The diagnostic moves from a \(1/1024\) one-shot reach regime to perfect
held-out realisation, splits an exactly public-equivalent hidden pair,
and removes sealed decision error.

Learning changes not only what can be inferred from evidence, but
sometimes what evidence can effectively be brought within reach. Making
that coordinate explicit turns a hidden assumption of evaluation into a
scientific question.

That question is larger than the construction but no less concrete. For
any claimed instance, we can ask what primitives were fixed, what budget
defined effective realisation, which observation law changed, which
worlds became distinguishable, and whether the distinction altered a
sealed decision. These questions are architecture-independent and
complement mature theories of experiment choice, information cost,
exploration, and skill.

A science of epistemic reach begins with a change of measurement. Alongside
accuracy, loss, and sample efficiency, we should measure the experiment
capability from which those outcomes are produced. Systems that appear
equal on today's evidence may differ in whether experience teaches them
how to obtain tomorrow's decisive evidence. Wide Learning studies how
experience changes which task-relevant evidence can reliably enter
computation under bounded resources.

\section*{Code and materials
availability}\label{code-and-materials-availability}
\addcontentsline{toc}{section}{Code and materials availability}

The construction uses no external dataset. Appendix A specifies the
reference implementation and exact reproduction command. The reference
implementation, configuration, validity checks, and one-command
reproduction script are available at
\url{https://github.com/chenjunzhou/Wide-Learning} (release
\texttt{paper-01-arxiv-submission-v1}).

\begin{center}\rule{0.5\linewidth}{0.5pt}\end{center}

\section{Appendix A. Reference construction and
implementation}\label{appendix-a.-reference-construction-and-implementation}

The reference implementation uses exact arithmetic over \(\mathbb F_2\)
and deterministic, namespaced seeds. It generates 20 seed-indexed
meta-environments, each with an independently sampled invertible affine
map. Every map contains 100 matrix entries and 10 bias bits. Sixteen
calibration pairs give the augmented design rank 11 and identify these
parameters exactly; 64 non-overlapping descriptors are held out for
deployment. Both \(\theta\) values are evaluated for every held-out
descriptor. The learner recovers each output bit by solving the
rank-11 linear system. At deployment it computes one address, consumes
one self-test attempt, probes only that address, and commits one binary
patch.

The public-equivalence certificate combines a construction-level proof
with exhaustive semantic checks. It parses the source of
\texttt{descriptor}, \texttt{api\_call}, \texttt{selftest}, and the
public response function, finding no read of the hidden mechanism. It
then compares paired public traces over all 1,024 descriptors for every
fixed meta-seed. The descriptor-leakage certificate checks the balanced
paired generator. The reach certificate reports the analytic \(q/2^d\)
bound as primary and labels Monte Carlo as secondary.

Raw core data contain 10,240 rows: four arms times 20 maps times 64
held-out descriptors times two mechanisms. Every row records the fixed
configuration, seed, descriptor, selected address, validity, probe
result, patch, truth, error, address attempts, nominal compute
iterations, and the identical primitive signature. Tables and reports
are regenerated from these rows.

With the released configuration and deterministic seed namespace, the
reference run reproduces the reported finite-run endpoints exactly. Runs
under alternative seeds constitute replication runs: their finite-sample
pre-learning hit rate and sealed error may vary, while the analytic
\(1/1024\) reach bound, exact public equivalence, and exact post-learning
recovery remain unchanged under the construction's assumptions.

After installing the package as described in the public repository
README, the full construction reproduces with the following commands
from the repository root:

\begin{Shaded}
\begin{Highlighting}[]
\BuiltInTok{cd}\NormalTok{ papers/01-learning-to-reach-evidence/code}
\ExtensionTok{bash}\NormalTok{ scripts/reproduce.sh}
\end{Highlighting}
\end{Shaded}

\section{Appendix B. Validity and alternative-explanation
checks}\label{appendix-b.-validity-and-alternative-explanation-checks}

{\def\LTcaptype{none} 
\begin{longtable}[]{@{}
  >{\raggedright\arraybackslash}p{(\linewidth - 6\tabcolsep) * \real{0.2308}}
  >{\raggedright\arraybackslash}p{(\linewidth - 6\tabcolsep) * \real{0.2308}}
  >{\centering\arraybackslash}p{(\linewidth - 6\tabcolsep) * \real{0.3077}}
  >{\raggedright\arraybackslash}p{(\linewidth - 6\tabcolsep) * \real{0.2308}}@{}}
\toprule\noalign{}
\begin{minipage}[b]{\linewidth}\raggedright
Code
\end{minipage} & \begin{minipage}[b]{\linewidth}\raggedright
Failure interpretation
\end{minipage} & \begin{minipage}[b]{\linewidth}\centering
Status
\end{minipage} & \begin{minipage}[b]{\linewidth}\raggedright
Decisive check
\end{minipage} \\
\midrule\noalign{}
\endhead
\bottomrule\noalign{}
\endlastfoot
F1 & Public-channel leak & PASS & Exact public-law certificate,
\(TV=0\) \\
F2 & Deployment search sufficient & PASS & Analytic \(1/1024<0.95\)
bound \\
F3 & Extra-affordance confusion & PASS & Identical five-primitive
signature in every arm \\
F4 & Test-specific lookup artefact & PASS & Fixed-repertoire valid rate
\(0\) \\
F5 & Compute-only explanation & PASS & A/B nominal compute exceeds C;
public law remains insufficient \\
F6 & Validator or label leak & PASS & Self-test is
\(\theta\)-independent; truth evaluator-only after patch \\
F7 & Memorisation artefact & PASS & Fresh held-out descriptors and 20
independent maps \\
F8 & Generic plasticity only & PASS & Separate fixed-evidence adaptation
succeeds without reach change \\
F9 & Epistemically irrelevant tool & PASS & Learned valid probe has
paired-world \(TV=1\) \\
F10 & No decision consequence & PASS & Sealed error falls to \(0\) \\
\end{longtable}
}

{\small
\bibliographystyle{unsrtnat}
\bibliography{WIDE_LEARNING_ARXIV}
}

\end{document}